\documentclass[11pt,letter]{article}

\usepackage{naacl2001}
\usepackage{macros}
\usepackage{multicol}  
\usepackage{amssymb}   
\usepackage{isolatin1} 

\include{epsf}

\title{A Complete \WN1.5 to \WN1.6 Mapping}
\author{J. Daud\'e, L. Padr\'o \and G. Rigau\\
        TALP Research Center\\
        Departament de Llenguatges i Sistemes Inform\`atics\\
        Universitat Polit\`ecnica de Catalunya. Barcelona\\
        {\tt \{daude,padro,g.rigau\}@lsi.upc.es} }

\begin{document}
\maketitle
\begin{abstract}We describe a robust approach for linking already existing 
lexical/semantic hierarchies. We use a constraint satisfaction 
algorithm (relaxation labelling) to select --among a set of
 candidates-- the node in a target taxonomy that bests matches 
each node in a source taxonomy. In this paper we present the 
complete mapping of the nominal, verbal, adjectival and
adverbial parts of WordNet 1.5 onto WordNet 1.6.
\end{abstract}

\section{Introduction}

There is an increasing need of having available general, accurate and
broad coverage multilingual lexical/semantic resources for developing
{\sc nl} applications. Thus, a very active field inside {\sc nl}
during the last years has been the fast development of generic
language resources.

Several attempts have been performed to connect already existing
ontologies.  For instance, to (semi)automatically link Spanish 
taxonomies extracted from {\sc dgile} \cite{dgile87} to their
English analogous from {\sc ldoce} \cite{ageno94}, or to WordNet
\cite{miller91} synsets \cite{rigau95};
the construction of large multilingual lexicons as in
\cite{knight94,okumura94}; or the alignment of {\sc edr} and
WordNet ontologies \cite{utiyama97}. Also, several lexical
resources and techniques are combined in \cite{atserias97,farreres98}
to map Spanish words from a bilingual dictionary to WordNet .

The use of relaxation labelling algorithm to attach substantial
fragments of the Spanish taxonomy derived from {\sc dgile}
\cite{rigau98} to the English WordNet has been reported in
\cite{daude99b}. The mapping of the nominal
parts of \WN1.5 to \WN1.6 is presented in \cite{daude00}.

In this paper we use the same approach to map the verbal, adjectival
and adverbial parts of \WN1.5 to \WN1.6, and present some improvements
on the results presented in \cite{daude00}. We provide evaluation
of the mapping accuracy by means of hand checking a random sample
of the taxonomies.

This paper is organized as follows: In section \ref{application} we
describe the used technique (the relaxation labelling algorithm) and
its application to hierarchy mapping. In section \ref{constraints} we
describe the constraints used in the relaxation process, and finally,
section~\ref{experiments} details performed tests and evaluation of
achieved results.

\section{Application of Relaxation Labelling to {\sc nlp}}
\label{application}

Relaxation labelling ({\sc rl}) is a generic name for a family of
iterative algorithms which perform function optimization, based on
local information, but with global effects.  See \cite{Torras89} for 
a summary.  Its most remarkable feature is that it can deal with any 
kind of constraints, and the algorithm is independent of the 
complexity of the model.

The algorithm has been applied to {\sc pos} tagging \cite{Marquez97a},
shallow parsing \cite{Voutilainen97} and to word sense disambiguation
\cite{Padro98a}.

\subsection{Algorithm Description}

The Relaxation Labelling algorithm deals with a set of variables (which
may represent words, synsets, etc.), each of which may take one among
several different labels ({\sc pos} tags, senses, {\sc mrd} entries,
etc.).  There is also a set of constraints which state compatibility
or incompatibility of a combination of pairs variable--label.

The aim of the algorithm is to find a weight assignment for each
possible label for each variable, such that (a) the weights for the
labels of the same variable add up to one, and (b) the weight
assignment satisfies --to the maximum possible extent-- the set of
constraints.

Summarizing, the algorithm performs constraint satisfaction to solve a
consistent labelling problem. The followed steps are:

\begin{enumerate}
\item Start with a random weight assignment.
\item Compute the {\sl support} value for each label of each variable.
  Support is computed according to the constraint set and to the
  current weights for labels belonging to context variables.
\item Increase the weights of the labels more compatible with the
  context (larger support) and decrease those of the less compatible
  labels (smaller support). Weights are changed proportionally to the
  support received from the context.
\item If a stopping/convergence criterion is satisfied, stop,
  otherwise go to step 2. We use the criterion of stopping when 
  there are no more changes, although stopping criteria may range
  from simply a fixed number of iterations to more sophisticated heuristic
  procedures \cite{Eklundh78}.
\end{enumerate}

\subsection{Application to taxonomy mapping}

As described in previous sections, the problem we are dealing with is
to map two taxonomies. In this particular case, we are interested in
mapping \WN1.5 to \WN1.6, that is, assign each synset in the former to
at least one synset in the later.

The modeling of the problem is the following:
\begin{itemize}
\item Each \WN1.5 synset is a variable for the relaxation
  algorithm. We will refer to it as {\sl source synset} and to
  \WN1.5 as {\sl source taxonomy}.
\item The possible labels for that variable are all the \WN1.6
  synsets which contain a word belonging to the source synset. We
  will refer to them as {\sl target synsets} and to \WN1.6 as 
  {\sl target taxonomy}.
\item The algorithm will need constraints stating whether a \WN1.6
  synset is a suitable assignment for a \WN1.5 synset.  As
  described in section~\ref{constraints}, these constraints will
  rely mainly on the taxonomy structure.
\end{itemize}

\section{The Constraints}
\label{constraints}

Constraints are used by the relaxation labelling algorithm to increase
or decrease the weight for a variable label. In our case, constraints
increase the weights for the connections between a source synset and a
target synset. Increasing the weight for a connection implies
decreasing the weights for all the other possible connections for the
same source synset.  To increase the weight for a connection,
constraints take into account already connected nodes that have the same
relationships in both taxonomies.

\subsection{Constraints for \WN\ Nouns}

Although there is a wide range of relationships between WordNet
synsets which can be used to build constraints, in \cite{daude00} 
we focus on the hyper/hyponym relationships, increasing the weight for a
connection when the involved nodes have hypernyms/hyponyms also
connected. Hyper/hyponym relationships are considered either directly
or indirectly (i.e. ancestors or descendants), depending on the kind
of constraint used.

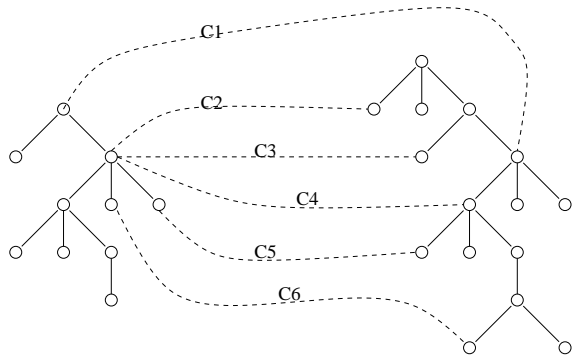
\figeps{exemple1}{Example of connections between taxonomies.}{f-exemple1}

Figure~\ref{f-exemple1} shows an example of possible connections
between two taxonomies. Connection $C_4$ will have its weight
increased due to $C_5$, $C_6$ and $C_1$, while connections $C_2$ and
$C_3$ will have their weights decreased.


In \cite{daude00} we distinguish different kinds of constraints, depending
on whether hyponyms, hypernyms or both are taken into account, on 
whether those relationships are considered direct or indirect, and on 
which of both taxonomies recursion is allowed.  Each constraint may be used
alone or combined with others.

Thus, the constraints used in \cite{daude00} are:

\begin{description}
\item[{\sc ii} constraints.]  This group includes {\sc iie}, {\sc iio}
  and {\sc iib} constraints.  All of them match an
  immediate--to--immediate relationship between both taxonomies, as
  shown in figure~\ref{f-ii}, where the arrows indicate an immediate
  hypernymy relationship. The nodes on the left hand side correspond
  to the source taxonomy and the nodes on the right to the target
  hierarchy. The dotted line is the connection which weight will be
  increased due to the existence of the connection(s) indicated with a
  continuous line.

\figeps{ii}{{\sc ii} constraints.}{f-ii}

\item[{\sc ai} constraints.]  This constraint group includes {\sc
    aie}, {\sc aio} and {\sc aib}.  That is, constraints that allow
  recursion on the source taxonomy.  A graphical representation is
  shown in figure~\ref{f-ai}, where the $+$ sign indicates that the
  hypernymy relationship represented by the arrow does not need to be
  immediate. In this case, this iteration is only allowed in the
  source taxonomy.

\figeps{ai}{{\sc ai} constraints.}{f-ai}

\item[{\sc ia} constraints:] Are symmetrical to {\sc ai} constraints.
  In this case, recursion is allowed only on the target taxonomy.

\item[{\sc aa} constraints:] Include the same combinations than above,
but allowing recursion on both sides, as presented in figure~\ref{f-aa}.
\end{description}

\figeps{aa}{{\sc aa} constraints.}{f-aa}

\subsection{Generalized Constraints}
\label{generalized}

Although the minimalist approach described above 
yields very good results for the WordNet nominal
taxonomy using only hyper/hyponymy relationships, it 
becomes much less useful in the case of verbs --due to the
very flat verb taxonomy in \WN-- and completely useless 
for adjectives and adverbs, which do not have such 
a kind of relationships.

Thus, we generalized the structural matching idea to relationships 
other than hyper/hyponyms. In that way, we can use constraints 
that involve any \WN\ relationship such as antonymy, ``also-see'', 
``similar-to'', etc.

This generalized constraint class has the schema presented
in figure~\ref{f-rr}, where a connection between a source synset
$s_1$ and a target synset $t_1$ is reinforced if there are a source
synset $s2$ and a target synset $t2$ such that $s_1 {\cal R} s_2$ and 
$t_1 {\cal R} t_2$, where ${\cal R}$ may be any \WN\ relationship.

\figeps{rr}{Generalized constraints.}{f-rr}

This new scenario enables us to map the complete \WN1.5 to \WN1.6.
Now we can talk about source and target {\sl graphs}, since it
is not only hyper/hyponymy what defines the structure which
constraints the mapping.

Nevertheless, this graph mapping is performed incrementally
for efficiency reasons, as  described in 
section \ref{incremental}.

\subsection{Additional Heuristic Constraints}

Apart from the generalized structural constraints presented
in section~\ref{generalized}, some heuristic, non-structural
constraints were also used to help the algorithm decide
in some cases in which structure was not enough informative.
For instance, the case where a leaf node in \WN1.5 may 
map to two different leafs in \WN1.6, both under the same
parent, may be disambiguated successfully with some simple 
heuristic constraints. 

A rough measure of similarity between two synsets can 
be computed using the number of word coincidences, the number of
non--empty word coincidences in glosses, or --in the case
of verb synsets-- the number of frame coincidences.

For each of these similarity measures, a non-structural constraint 
can be stated, which reinforces a connection
between a source and a target synset proportionally to
their similarity.

We denote these constraints with {\sc w}, {\sc g} and {\sc f},
depending on whether word, gloss or frame coincidences
are used to compute similarity.

\subsection{Constraints for \WN\ Verbs, Adjectives and Adverbs}
\label{incremental}

  In order to reduce the computation time for the mapping,
we proceeded in three phases, following the relationships that
establish dependences among different {\sc pos} in WordNet, as described below:

\begin{itemize}
\item  First, we mapped nouns, using only hyper/hyponymy relationships
  plus {\sc w} and {\sc g} additional constraints. 
  
  Similarly, verbs were mapped using  hyper/hyponymy, antonymy and the
  ``also-see'' \WN\ relationship. {\sc w}, {\sc g} and {\sc f} 
  additional constraints were also used.
  
\item  Second, adjectives were mapped using {\sl adj--to--adj} relationships
  such as antonymy, ``similar-to'' and ``also-see'', as well as the 
  {\sl adj--to--verb} relationship ``participle-of'' and the 
  {\sl adj--to--noun} ``pertains''  and ``attribute''. 
  {\sc w} and {\sc g} additional constraints were also used.
  
  Notice that the graph used to check the 
  constraints imposed by {\sl adj--to--verb} and {\sl adj--to--noun}
  relationships
  was the result of the previous steps. That means that verbs and nouns
  are already mapped, reducing considerably the search space and 
  accelerating relaxation labelling convergence. Since the noun and
  verb mappings do not depend on adjectives (there are no {\sl noun--to--adj}
  relationships and only one {\sl verb--to--adj} --not used to avoid 
  circularity--), results wouldn't have been greatly affected if the 
  mappings had been performed in parallel, but the convergence would 
  have been slower.
  
\item The third and last phase is the adverb mapping, which is
  performed with the only {\sl adv--to--adv} relationship, antonymy and
  with the {\sl adv--to--adj} ``derived''. {\sc w} and {\sc g} additional 
  constraints were also used.
  
  Again, the noun, verb, and adjective graphs are taken as a  
  static picture of the mapping obtained from previous phases.
\end{itemize}

\section{Experiments and Results}
\label{experiments}

In the performed tests we evaluated the performance of different
constraint combinations for each phase. 

All figures presented in this section were computed by
manually linking to \WN1.6 a sample chosen
from \WN1.5, and then use this sample mapping as a reference.
 The validation sample consists of 1900 noun
synsets, 1000 verb, 1000 adjective and 300 adverb synsets.

 See \cite{daude00} for a comparison of our mapping with the {\sl SenseMap} provided by Princeton\footnote{See {\sc wn} web page at \\
                   http://www.cogsci.princeton.edu/\ona wn/}.

Results are presented in tables~\ref{taula-b}, \ref{taula-bwg}  
and \ref{taula-tot}. They present the results for increasingly
complex constraint sets.

First, table~\ref{taula-b} presents results with
the {\sl basic} constraint set for each part--of--speech. They
consist of the hyper/hyponym relationships (when available) plus
other available {\sl intra-}{\sc pos} relationships --i.e. relationships
between synsets with the same {\sc pos}. Namely, the constraint sets are:
\begin{description}
\item[Nouns:] {\sc aa} hyper/hyponym constraint set --corresponding
to the basic technique used in \cite{daude00}.
\item[Verbs:] {\sc aa} hyper/hyponym constraint set 
plus ``also-see'' and antonymy relationships in {\sc ii} 
(immediate--to--immediate) mode.
\item[Adjectives:] Antonymy, ``similar-to'' and ``also-see'', all of
them in {\sc ii} mode.
\item[Adverbs:] Antonymy in {\sc ii} mode.
\end{description}

\begin{table}[htb] \centering
\begin{tabular}{l|c|c|c|c|}
    &  Cover. & ambiguous.      &  overall \\ \hline 
 N  &   99.7\% & 94.9\%--99.6\% &  97.6\%--99.8\% \\ \hline 
 V  &   96.9\% & 93.5\%--99.2\% &  94.6\%--99.2\% \\ \hline 
 A  &   94.1\% & 82.8\%--98.9\% &  89.5\%--99.4\% \\ \hline 
 R  &   80.8\% & 97.5\%--100.0\% &  99.0\%--100.0\% \\ \hline 
\end{tabular}
\caption{Precision--recall results for {\sl basic} constraint set.}
\label{taula-b}
\end{table}

Second, table~\ref{taula-bwg} shows the results for the {\sl basic}
set extended with the {\sc w} and {\sc g} constraints for
nouns, adjectives and adverbs, and with
{\sc w}, {\sc g} and {\sc f} for verbs.

Note that the results for verbs are lower over all
synsets than over ambiguous ones. This is caused because
some of the non--ambiguous \WN1.5 synsets in the validation 
sample were hand marked to have no correspondence in \WN1.6 
(i.e. they were removed in the new version), which causes
the precision for non--ambiguous nodes to be only 99.1\%.

\begin{table}[htb] \centering
\begin{tabular}{l|c|c|c|c|}
    &  Cover. & ambiguous.      &  overall \\ \hline 
 N  &   99.9\% & 97.5\%--97.7\% &  98.8\%--98.9\% \\ \hline 
 V  &   99.8\% & 99.4\%--99.7\% &  99.3\%--99.6\% \\ \hline 
 A  &   98.9\% & 96.5\%--98.8\% &  97.9\%--99.3\% \\ \hline 
 R  &   99.5\% & 97.5\%--100.0\% &  99.0\%--100.0\% \\ \hline 
\end{tabular}
\caption{Precision--recall for {\sl basic}+{\sc wgf} constraint set}
\label{taula-bwg}
\end{table}

Third, table~\ref{taula-be} shows the results for the {\sl basic}
set extended with available {\sl extra}--{\sc pos} relationships, which
are:
\begin{description}
\item[Nouns:] No {\sl extra}--{\sc pos} relationships.
\item[Verbs:] No {\sl extra}--{\sc pos} relationships.
\item[Adjectives:] ``participle-of'', ``attribute'' ({\sl adj--to--noun}) and 
``pertains-to'' ({\sl adj--to--verb}), both in {\sc ii} mode.
\item[Adverbs:] ``derived--from'' ({\sl adv--to--adj}), in {\sc ii} mode.
\end{description}

\begin{table}[htb] \centering
\begin{tabular}{l|c|c|c|c|}
    &  Cover. & ambiguous.      &  overall \\ \hline 
 N  &    --    &       --       &        --       \\ \hline 
 V  &    --    &       --       &        --       \\ \hline 
 A  &   95.8\% & 85.0\%--98.9\% &  90.9\%--99.4\% \\ \hline 
 R  &   88.0\% & 69.2\%--94.2\% &  87.9\%--98.1\% \\ \hline 
\end{tabular}
\caption{Precision--recall for {\sl basic}+{\sl extra}--{\sc pos} constraint set}
\label{taula-be}
\end{table}

Finally, Table~\ref{taula-tot} shows the results obtained when
using together the {\sl basic} set, plus the {\sc wgf} set, plus the
{\sl extra}-{\sc pos} set. For nouns and verbs, results from
table~\ref{taula-bwg} are repeated for clarity.

\begin{table}[htb] \centering
\begin{tabular}{l|c|c|c|c|}
    &  Cover. & ambiguous.      &  overall \\ \hline 
 N  &   99.9\% & 97.5\%--97.7\% &  98.8\%--98.9\% \\ \hline 
 V  &   99.8\% & 99.4\%--99.7\% &  99.3\%--99.6\% \\ \hline 
 A  &   99.0\% & 96.5\%--99.1\% &  97.9\%--99.5\% \\ \hline 
 R  &   99.6\% & 98.3\%--100.0\% &  99.3\%--100.0\% \\ \hline 
\end{tabular}
\caption{Precision--recall for complete constraint set.}
\label{taula-tot}
\end{table}

  Those figures indicate that in all cases, the use of
additional constraints other than the {\sl basic} set, produces an
increase in coverage. 

For the {\sc wgf} constraints, this improvement
is accompanied by a large ambiguity reduction (thus a narrower
precision--recall interval) and an increment in precision,
specially in verbs and adjectives. Adverbs increase 
coverage maintaining precision and recall, while nouns present 
a small improvement.

The {\sl extra}-{\sc pos} constraints turned to be useful for adjectives, 
but harmful for adverbs when used alone. In both cases, they
raised precision when used in conjunction with {\sc wgf} constraints.

The recall below 100\% is caused by synsets having several candidates
with the same structural features. Since all candidates obtain
the same support from the constraints, the algorithm cannot chose one
or another and leaves it ambiguous. If the final support is under the
threshold, no answer is computed as correct.

The precision under 100\% is caused by two kinds of errors:
First, some \WN1.5 synsets are splitted in \WN1.6 and the human
evaluator may consider that the best choice is not the one
made by the system. Second, some \WN1.5 synsets simply do not
appear in \WN1.6, but if the system assigns them anyway,
they are computed as an error.

\section{Conclusions \& Further Work}

We have applied the relaxation labelling algorithm to build a
complete mapping between two different WordNet versions, and
reported results on \WN1.5 to \WN1.6 complete 
mapping\footnote{The mapping has been made public at 
          http://www.lsi.upc.es/\ona nlp.
          So far, 37 licenses have been provided.}.

An incremental approach has been used to map separately the different
\WN\ {\sc pos} files, using at each step the results of previous ones.

Non--structural information such as synset words, glosses and
verb frames has been proven useful in increasing coverage, reducing
ambiguity, and improving performance.

\section{Acknowledgments}

This research has been partially funded by the the UE Commission 
(NAMIC IST-1999-12392), by the Spanish Research Department 
(TIC98-423-C06-06, TIC2000-1735-C02-02) and by
the Catalan Research Department, (Quality Research 
Group Programme, GRQ93-3015).

\bibliographystyle{acl}
\bibliography{/usr/usuaris/ia/padro/articles/fullbib}

\end{document}